\documentclass[12pt,twocolumn]{article}
\usepackage{amsmath}
\usepackage{amssymb}
\usepackage{amsthm}
\usepackage{epsfig}

\title{\vspace*{-.3in}\Large \bf Binary Classification Based on Potential Functions}

\author{\bf Erik Boczko\footnote{
Biomedical Informatics,
Vanderbilt University,
Nashville, TN 37232}
\and
\bf Andrew Di Lullo\footnote{Undergraduate student,
Department of Physics,
Ohio University,
Athens, OH 45701}
\and
\bf Todd Young\footnote{Corresponding author,
Department of Math,
Ohio University,
Athens, OH 45701}
}

\newcommand{\D}[1]{{\mathbb#1}}

\newcommand{\RR}{{\D{R}}}

\newcommand{\xb}{\mathbf{x}}
\newcommand{\yb}{\mathbf{y}}
\newcommand{\zb}{\mathbf{z}}
\newcommand{\ab}{\mathbf{a}}
\newcommand{\bb}{\mathbf{b}}
\newcommand{\cb}{\mathbf{c}}
\newcommand{\Cb}{\mathbf{C}}

\begin{document}

\date{}

\maketitle

\thispagestyle{empty}

\noindent
{\em
Abstract --
We introduce a simple and computationally trivial method
for binary classification based on the evaluation of
potential functions. We demonstrate that despite the
conceptual and computational simplicity of the method
its performance can match or exceed that of standard Support
Vector Machine methods.
}

\vskip .2cm

\noindent
{\bf Keywords: Machine Learning, Microarray Data}

\bigskip
\noindent
{\large \bf 1 \ Introduction}
\medskip

\noindent
Binary classification is a fundamental focus in machine learning and
informatics with many possible applications. For instance in biomedicine,
the introduction of microarray and proteomics data has opened the door to
connecting a molecular snapshot of an individual with the presence or absence
of a disease. However, microarray data sets can contain tens to hundreds
of thousands
of observations and are well known to be noisy~\cite{brody}.
Despite this complexity, algorithms exist that are capable
of producing very good performance~\cite{stat1,stat2}.
Most notable among these methods are the Support Vector Machine (SVM) methods.
In this paper we introduce a simple and computationally trivial method
for binary classification based on potential functions. This classifier,
which we will call the {\it potential method}, is in a
sense a generalization of the nearest neighbor methods and is also
related to radial basis function networks (RBFN)~\cite{krzyzak}, another
method of current interest in machine learning. Further, the method can be
viewed as one possible nonlinear version of Distance Weighted Discrimination (DWD),
a recently proposed method whose linear version consists of choosing a decision plane by
minimizing the sum of the inverse distances to the plane \cite{RMW}.

Suppose that $\{\yb_i\}_{i=1}^m$ is a set of data of one type, that
we will call {\em positive} and $\{\zb_i\}_{i=1}^n$ is a data set
of another type that we call {\em negative}. Suppose that
both sets of data are vectors in $\RR^N$. We will assume that
$\RR^N$ decomposes into two sets $Y$ and $Z$ such that
each $\yb_i \in Y$, $\zb_i \in Z$ and any point in $Y$ should
be classified as positive and any point in $Z$ should be classified as negative.
Suppose that $\xb \in \RR^N$ and we wish to predict whether $\xb$ belongs
to $Y$ or $Z$ using only information from the finite sets of data
$\{\yb_i\}$ and $\{\zb_i\}$. Given distance functions $d_1(\cdot,\cdot)$
and $d_2(\cdot,\cdot)$ and positive constants $\{a_i\}_{i=1}^m$, $\{b_i\}_{i=1}^n$,
$\alpha$ and $\beta$ we define a potential function:
\begin{equation}\label{pot}
    I(\xb) =  \sum_{i=1}^m \frac{a_i}{d_1(\xb,\yb_i)^\alpha}
                    -  \sum_{i=1}^n \frac{b_i}{d_2(\xb,\zb_i)^\beta}.
\end{equation}
If $I(\xb) > 0$ then we say that $I$ classifies $\xb$ as belonging to
$Y$ and if $I(\xb)$ is negative then $\xb$
is classified as part of $Z$. The set $I(\xb) = 0$ we call the decision
surface. Under optimal circumstances it should coincide with the boundary
between $Y$ and $Z$.

Provided that $d_1$ and $d_2$ are sufficiently easy to evaluate, then
evaluating $I(\xb)$ is computationally trivial. This fact could make
it possible to use the training data to search for optimal choices of
$\{a_i\}_{i=1}^m$, $\{b_i\}_{i=1}^n$, $\alpha$, $\beta$ and even the
distance functions $d_j$. An obvious choice for $d_1$ and $d_2$ is
the Euclidean distance. More generally, $d$ could be chosen as the distance defined
by the $\ell^p$ norm, i.e.\ $d(\xb,\yb) = \| \xb - \yb \|_p$ where
\begin{equation}\label{norm}
 \| \xb \|_p \equiv \left( x_1^p + x_2^p + \ldots + x_N^p \right)^{1/p}.
\end{equation}
A more elaborate choice for a distance $d$ might be the following.
Let $\cb = (c_1, c_2, \ldots, c_N)$ be an $N$-vector and define $d_\cb$
to be the $\cb$-weighted distance:
\begin{multline}\label{dc}
  d_{\cb,p}(\xb,\yb) \equiv \left( c_1|x_1 - y_1|^p + c_2|x_2-y_2|^p \right. \\ + \ldots +
                               \left. c_N|x_N - y_N|^p \right)^{1/p}.
\end{multline}
This distance allows assignment of different weights to the various attributes.
Many methods for choosing $\cb$ might be suggested and we propose a few here.
Let $\Cb$ be the vector associated with the classification of the data points,
$C_i = \pm 1$ depending on the classification of the i-th data point.
The vector $\cb$ might consist of the absolute values univariate c
orrelation coefficients associated with the $N$ variables with respect
to $\Cb$. This would have the effect of emphasizing directions which
should be emphasized, but very well might also suppress directions
which are important for multi-variable effects. Choosing $\cb$ to be
$1$ minus the univariate $p$-values associated with
each variable could be expected to have a similar effect.
Alternatively, $\cb$ might be derived from some
multi-dimensional statistical methods.
In our experiments it turns out that $1$ minus the $p$-values works
quite well.

Rather than $a_i = b_i =1$ we might consider other weightings of
training points. We would want to make the choice of
$\ab = (a_1, a_2, \ldots, a_m)$ and $\bb = (b_1, b_2, \ldots, b_n)$
based on easily available information. An
obvious choice is the set of distances to other test points. In
the checkerboard experiment below we demonstrate that training points
too close to the boundary between $Y$ and $Z$ have undue influence
and cause irregularity in the decision curve. We would like to
give less weight to these points by using the distance from the
points to the boundary. However, since the boundary is not known,
we use the distance to the closest point in the other set as an
approximation. We show that this approach gives improvement in
classification and in the smoothness of the decision surface.

Note that if $p=2$ in (\ref{norm})
our method limits onto the usual nearest neighbor method as
$\alpha = \beta \rightarrow \infty$ since for large $\alpha$ the term
with the smallest denominator will dominate the sum. For finite
$\alpha$ our method gives greater weight to nearby points.

In the following we report on tests of the efficacy of the
method using various $\ell^p$ norms as the distance, various choices of
$\alpha = \beta$ and a few simple choices for $\cb$,
$\ab$, and $\bb$.

\bigskip
\noindent
{\large \bf 2 \ A Simple Test Model}
\medskip

\begin{figure}[hbtp]
\centerline{\hbox{\psfig{figure=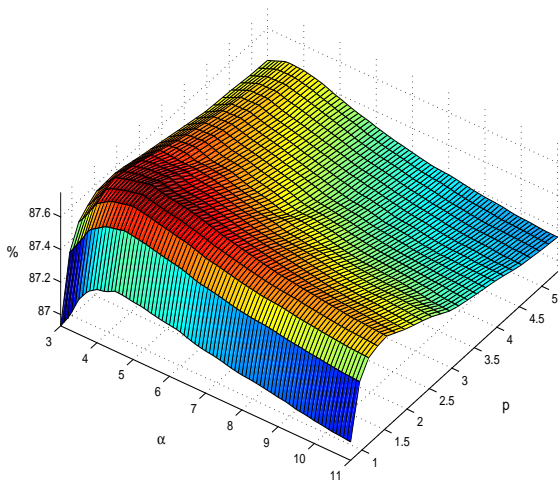,height=7cm,width=7.5cm}}}
\caption{\small The percentage of correct classifications for the
4 by 4 checkerboard test problem as a function of the parameters
$\alpha$ and $p$. The maximum occurs near $p=1.5$ and $\alpha = 4.5$.
Notice that the graph is fairly flat near the maximum.}
\label{fig:p-alpha}
\end{figure}

\noindent
We applied the method to the model problem
of a 4 by 4 checkerboard. In this test we suppose that
a square is partitioned into a 16 equal subsquares and suppose
that points in alternate squares belong to two distinct
types. Following \cite{MM}, we used 1000 randomly selected points
as the training set and 40,000 grid points as the test set.
We choose to define both the distance functions by the usual $\ell^p$ norm.
We will also require $\alpha = \beta$ and $a_i = b_i  =1$
Thus we used as the potential function:
\begin{equation}\label{normpot}
    I(\xb) =  \sum_{i=1}^m \frac{1}{\|\xb-\yb_i\|_p^{\alpha}}
                    - \sum_{i=1}^n \frac{1}{\|\xb-\zb_i\|_p^{\alpha}}.
\end{equation}
Using different values of $\alpha$ and $p$ we found the percentage of
test points that are correctly classified by $I$. We repeated this
experiment on 50 different training sets and tabulated the percentage
of correct classifications as a function of $\alpha$ and $p$.
The results are displayed in Figure~\ref{fig:p-alpha}. We find
that the maximum occurs at approximately
$p = 1.5$ and $\alpha = 4.5$.

The relative flatness near the maximum in Figure~1
indicates robustness of the method with respect to these
parameters. We further observed that changing the training
set affects the location of the maximum only slightly and the affect on
the percentage correct is small.

\begin{figure}[h!]
\centerline{\psfig{figure=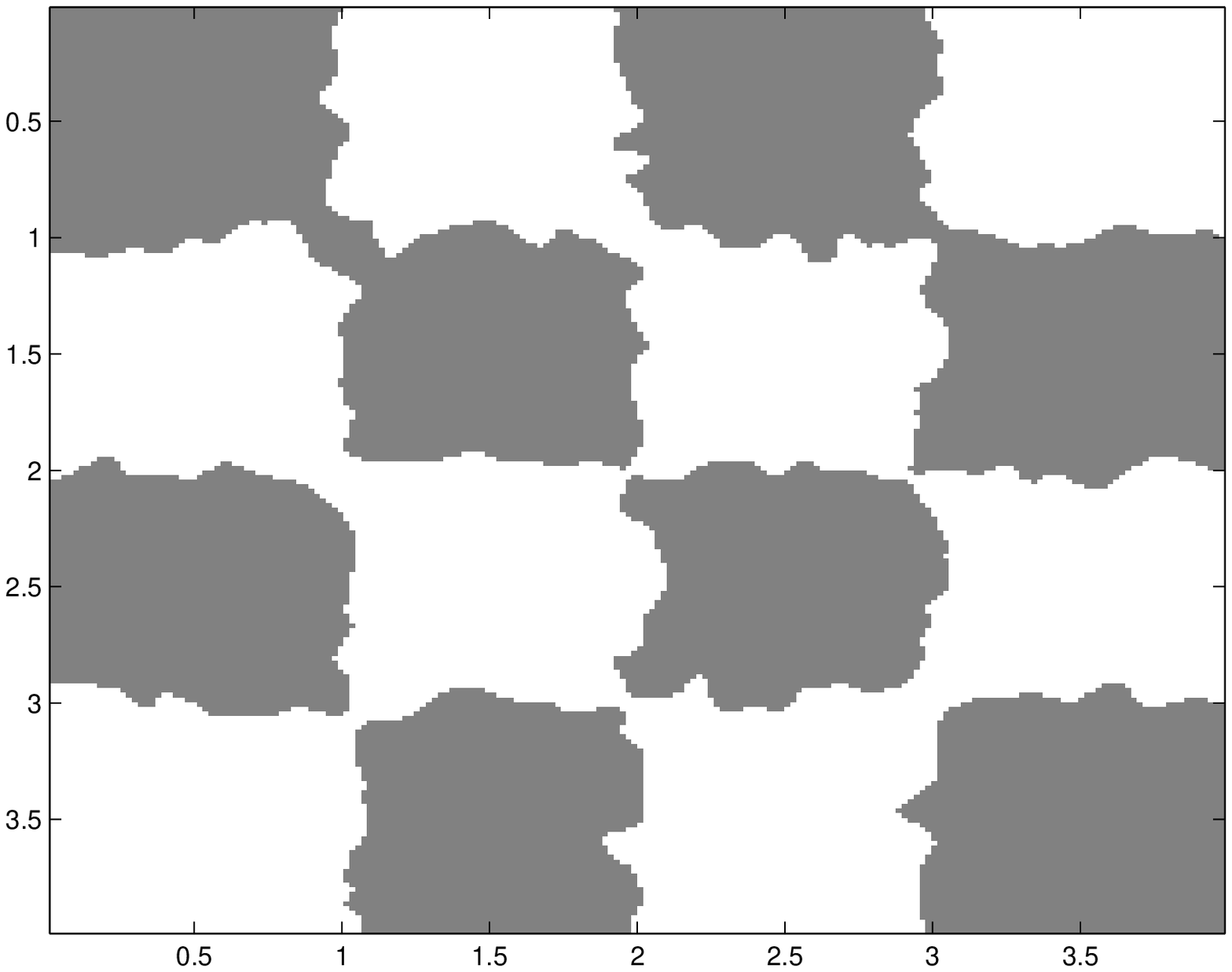,height=6cm,width=6cm}}
\centerline{\psfig{figure=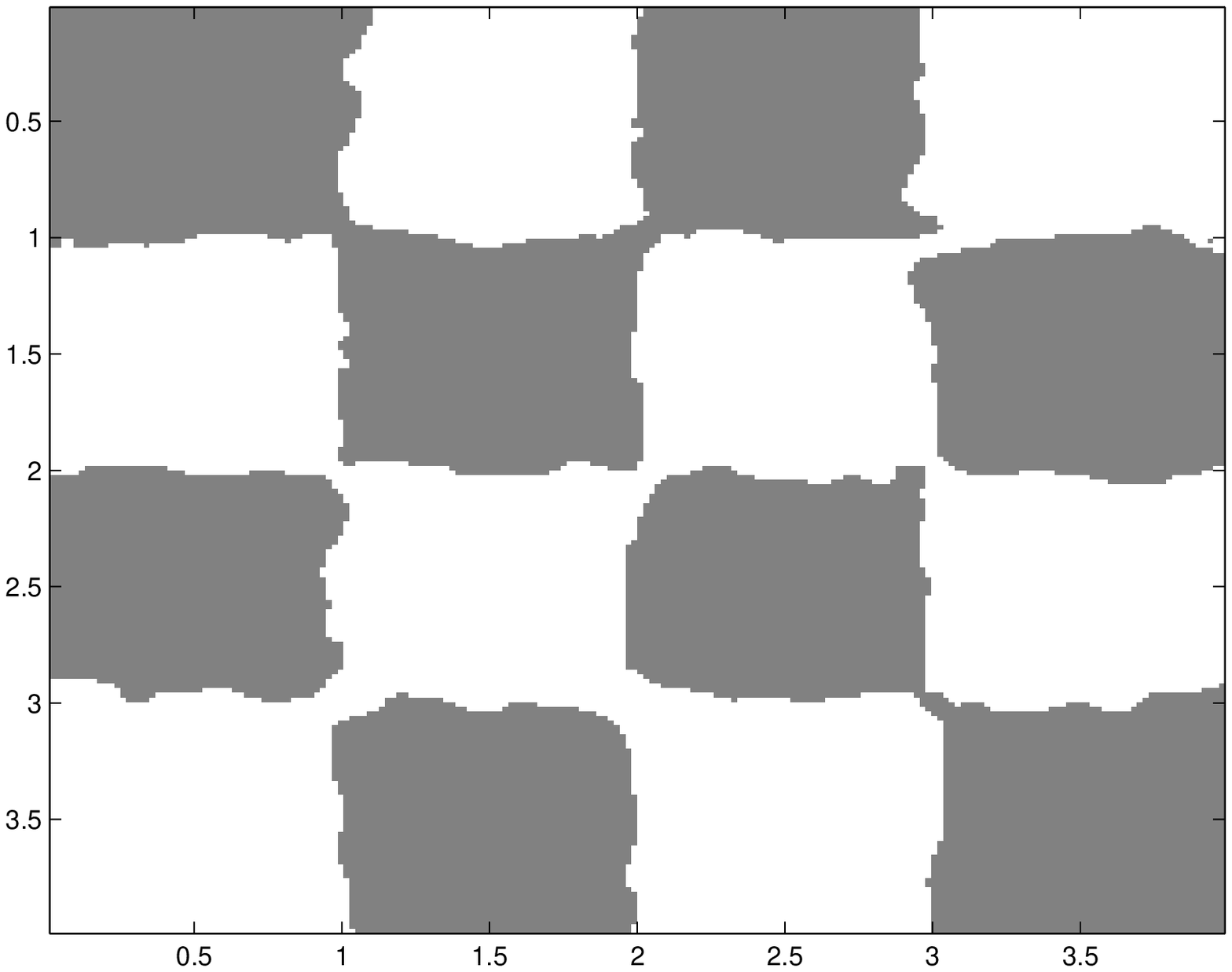,height=6cm,width=6cm}}
\caption{\small (a) The classification of the 4 by 4 checkerboard without
distance to boundary weights. In this test $95\%$ were correctly
classified.
(b) The classification using
distance to boundary weights.
Here $96.2\%$ were correctly classified.}
\label{fig:square}
\end{figure}

Finally, we tried classification of the 4 by 4 checkerboard using
the minimal distance to data of the opposite type in the coefficients
for the training data, i.e. $\ab$ and $\bb$ in:
$$
    I(\xb) =  \sum_{i=1}^m
       \frac{(1+\epsilon)a_i^\beta}{\|\xb-\yb_i\|_p^\alpha}
       - \sum_{i=1}^n
            \frac{(1-\epsilon)b_i^\beta}{\|\xb,\zb_i\|_p^\alpha}.
$$
With this we obtained
$96.2\%$ accuracy in the classification and a noticably smoother
decision surface (see Figure~\ref{fig:square}(b)).
The optimized parameters for our method were $p \approx 3.5$ and
$\alpha \approx 3.5$. In this optimization we also used the
distance to opposite type to a power $\beta$ and the optimal
value for $\beta$ was about $3.5$. In
\cite{MM} a SVM method obtained $97\%$ correct classification,
but only after 100,000 iterations.

\bigskip
\noindent
{\large \bf 3 \ Clinical Data Sets}
\medskip

\noindent
Next we applied the method to micro-array data from two cancer study
sets \verb&Prostate_Tumor& and \verb&DLBCL& \cite{stat1,stat2}.
Based on our experience in the previous problem, we used the potential function:
\begin{equation}\label{greatpot}
    I(\xb) =
 \sum_{i=1}^m \frac{(1+\epsilon)a_i^\beta}{d_{\cb,p}(\xb,\yb_i)^\alpha}
      - \sum_{i=1}^n \frac{(1-\epsilon)b_i^\beta}{d_{\cb,p}(\xb,\zb_i)^\alpha},
\end{equation}
where $d_{\cb,p}$ is the metric defined in (\ref{dc}). The vector
$c_i$ was taken to be $1$ minus the univariate p-value for each
variable with respect to the classification.
The weights $a_i$, $b_i$ were taken to be the distance from each
data point to the nearest data point of the opposite type.
Using the potential
(\ref{greatpot}) we obtained leave-one-out cross validation (LOOCV) for various
values of $p$, $\alpha$, $\beta$, and $\epsilon$. For these data sets
LOOCV has been shown to be a valid methodology \cite{stat1}

On the \verb&DLBCL& data the nearly optimal performance of $98.7\%$ was acheived
for many parameter combinations. The SVM methods studied in \cite{stat1,stat2}
achieved $97.5\%$ correct on this data while the $k$-nearest neightbor correctly
classified only $87\%$. Specifically, we found that for each
$1.6 \le p \le 2.4$ there were robust sets of parameter combinations
that produced performance better than SVM. These parameter sets were contained
generally in the intervals: $10 < \alpha < 15$ and $10 < \beta < 15$ and
$0 < \epsilon < .5$.

For the \verb&DLBCL& data when we used the $\ell^p$ norm instead of the
weighted distances and also dropped the data weights ($\epsilon = \beta = 0$)
the best performance sank to $94.8\%$ correct
classification at $(p,\alpha) = (2,6)$. This illustrates the
importance of these parameters.

For the \verb&Prostrate_tumor& data set the results using potential
(\ref{greatpot}) were not quite as good. The best
performance, $89.2\%$ correct, occured for $1.2 \le p \le 1.6$ with
$\alpha \in [11.5, 15$, $\beta \in [12,14]$, $\epsilon \in [.1,.175]$ .
In \cite{stat1,stat2} various SVM methods were shown to
achieve $92\%$ correct and the $k$-nearest neighbor method acheived $85\%$ correct.
With feature selection we were able to obtain much
better results on the \verb&Prostrate_tumor& data set. In particular,
we used the univariate $p$-values to select the most relevant features.
The optimal performance occured with 20 features.
In this test we obtain $96.1\%$ accuracy for a robust set of
parameter values.

\begin{table}[hbt]
\centerline{
\begin{tabular}{|c|cccc|}
\hline
      data set  &  kNN    &  SVM      & Pot        & Pot-FS    \\ \hline
        DLBCL   &  $87\%$ & $97.5\%$  & $98.7\%$   &    ---   \\
      Prostate  &  $85\%$ & $92\%$    & $89.2\%$   & $96.1\%$ \\
\hline
\end{tabular}
}
\vskip .2cm
\caption{\small Results from the potential method on benchmark DLBCL
and Prostate\_tumor micro-array data sets compared
with the SVM methods and the k-nearest neighbor method.
The last column is the performance of the potential method with univariate
feature selection.}
\label{cancer}
\end{table}

\bigskip
\noindent
{\large \bf 4 \ Conclusions}
\medskip

\noindent
The results demonstrate that, despite its simplicity,
the potential method can be as
effective as the SVM methods.  Further work needs
to be done to realize the maximal performance of the method.
It is important that most of the calculations required by the
potential method are mutually independent and so are highly parallelizable.

We point out an important difference between the potential method
and Radial Basis Function Networks. RBFNs were originally
designed to approximate a real-valued function on $\RR^N$. In classification
problems, the RBFN attempts to approximate the characteristic functions
of the sets $Y$ and $Z$ (see \cite{krzyzak}). A key point of our method
is to approximate the decision surface only. The
potential method is designed for classifcation problems whereas
RBFNs have many other applications in machine learning.

We also note that the potential method, by putting signularities at
the known data points, always classifies some neighborhood of a data
point as being in the class of that point. This feature
makes the potential method less suitable when the decision
surface is in fact not a surface, but a ``fuzzy" boundary region.

There are several avenues of investigation that seem to be worth
pursuing.  Among these, we have further investigated the role of the
distance to the boundary with success \cite{BY}. Another direction
of interest would be to explore alternative
choices for the weightings $\cb$, $\ab$ and $\bb$. Another
would be to investigate the use of more general metrics
by searching for optimal choices in a suitable function space
\cite{sda}. Implementation of feature selection with the potential
method is also likely to be fruitful. Feature selection routines
already exist in the context of $k$-nearest neighbor mathods
\cite{li} and those can be expected to work equally well for the
potential method. Feature selection is recongnized to be very
important in micro-array analysis,  and we view the success of the
method without feature selection and with primative feature selection
as a good sign.


\small
\begin{thebibliography}{99}

\bibitem{BY}
E.M.~Boczko and T.~Young,
Signed distance functions: A new tool in binary classification,
ArXiv preprint: CS.LG/0511105.

\bibitem{brody} J.P.~Brody, B.A.~Williams, B.J.~Wold and S.R.~Quake, Significance and statistical errors in the analysis of DNA microarray data.
{\em Proc.~Nat.~Acad.~Sci.}, {\bf 99} (2002), 12975-12978.

\bibitem{roc} D.H.~Hand, R.J.~Till, A simple generalization of the area under the ROC
curve for multiple class classification problems,
{\em Machine Learning.} {\bf 45} (2001), 171-186.
\bibitem{krzyzak} A.~Krzy$\dot{z}$ak, Nonlinear function learning using optimal
radial basis function networks, {\em Nonlinear Anal.}, {\bf 47}(2001), 293-302.

\bibitem{hopf} J.~J.~Hopfield, Neural networks and physical systems with emergent collective computational abilities,
{\em Proc. Natl. Acad. Sci.}, {\bf 79} (1982), 2554-2558.

\bibitem{li}
L.P.~Li, C.~Weinberg, T.~Darden, L.~Pedersen,
Gene selection for sample calssification based on gene
expresion data: study of sensitivity to choice of
parameters of the GA/KNN method, {\em Bioinformatics}
{\bf 17} (2001), 1131-1142.


\bibitem{MM} O.L.~Mangasarian and D.R.~Musicant, Lagrangian Support
Vector Machines, {\it J. Mach. Learn. Res.}, {\bf 1} (2001), no.~3, 161--177.



\bibitem{RMW} J.V.~Rogel, T.~Ma, M.D.~Wang,
Distance Weighted Discrimination and Signed Distance Function algorithms for
binary classification. A comparison study. Preprint, Georgia Institute of Technology, 2006.

\bibitem{sda} J.H.~Moore, J.S.~Parker, N.J.~Olsen,
Symbolic discriminant analysis of microarray data in autoimmune disease, {\em Genet. Epidemiol.} {\bf 23}
(2002), 57-69.

\bibitem{stat1}  A.~Statnikov, C.F.~Aliferis, I.~Tsamardinos, D.~Hardin, S.~Levy,
A Comprehensive Evaluation of Multicategory Classification Methods for
Microarray Gene Expression Cancer Diagnosis, {\it Bioinformatics} {\bf 21}(5), 631-43, 2005.

\bibitem{stat2} A.~Statnikov, C.F.~Aliferis, I.~Tsamardinos. Methods for Multi-Category Cancer Diagnosis
from Gene Expression Data: A Comprehensive Evaluation to Inform Decision Support System Development,
In {\it Proceedings of the 11th World Congress on Medical Informatics (MEDINFO)}, September 7-11, (2004),
San Francisco, California, USA




\end{thebibliography}
\end{document}